\def\year{2022}\relax
\documentclass[letterpaper]{article} 
\usepackage{aaai22}  
\usepackage{times}  
\usepackage{helvet}  
\usepackage{courier}  
\usepackage[hyphens]{url}  
\usepackage{graphicx} 
\usepackage{natbib}  
\usepackage{caption} 
\DeclareCaptionStyle{ruled}{labelfont=normalfont,labelsep=colon,strut=off} 
\usepackage{algorithm}
\usepackage{algorithmic}

\usepackage{newfloat}
\usepackage{listings}
\floatstyle{ruled}
\newfloat{listing}{tb}{lst}{}
\floatname{listing}{Listing}

\usepackage{multirow}
\usepackage{booktabs}
\title{IFBiD: Inference-Free Bias Detection}
\author {
    Ignacio Serna,
    Daniel DeAlcala,
    Aythami Morales, 
    Julian Fierrez, 
    Javier Ortega-Garcia 
}
\affiliations {
    Biometrics and Data Pattern Analytics Lab (BiDA-Lab), Autonomous University of Madrid\\
    \{ignacio.serna, daniel.dealcala, aythami.morales, julian.fierrez, javier.ortega\}@uam.es
    
}

\begin{document}

\maketitle

\begin{abstract}
   This paper is the first to explore an automatic way to detect bias in deep convolutional neural networks by simply looking at their weights, without the model inference for a specific input. Furthermore, it is also a step towards understanding neural networks and how they work. We analyze how bias is encoded in the weights of deep networks through a toy example using the Colored MNIST database and we also provide a realistic case study in gender detection from face images using state-of-the-art methods and experimental resources. To do so, we generated two databases with 36K and 48K  biased models each. In the MNIST models we were able to detect whether they presented strong or low bias with more than 99\% accuracy, and we were also able to classify between four levels of bias with more than 70\% accuracy. For the face models, we achieved 83\% accuracy in distinguishing between models biased towards Asian, Black, or Caucasian ethnicity.
\end{abstract}

\begin{figure}[t]
    \centering
    \includegraphics[width=0.91 \columnwidth]{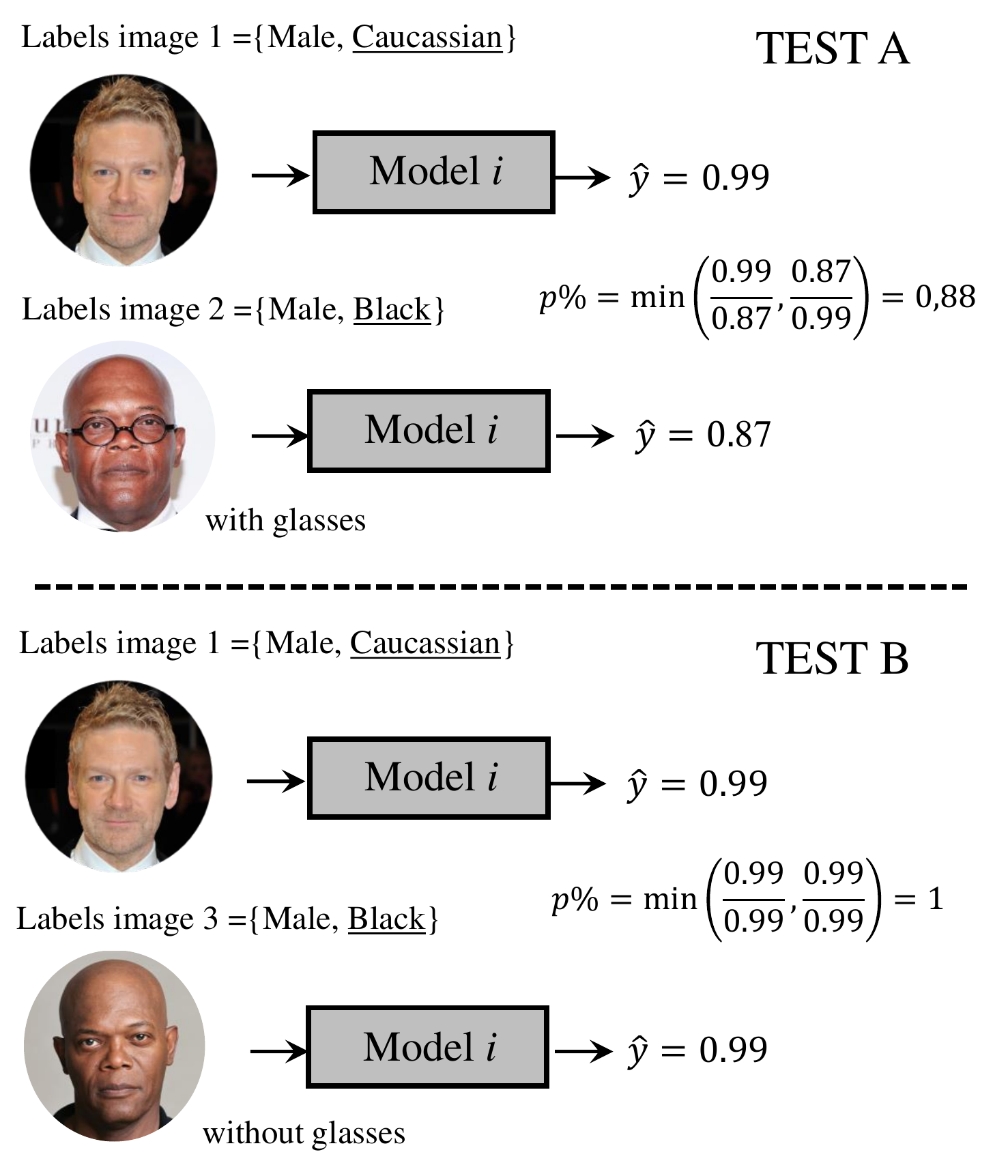}
    \caption{Traditional bias detection test based on inference analysis and the demographic parity measure $p\%$~\cite{zhang2018mitigating}. The Model $i$ is a gender classifier and the sensitive attribute $z$ is ethnicity. In this example, Test A suggests that the Model $i$ is biased with respect to ethnicity. Test B revealed that the difference in $p\%$ is caused not by ethnicity but by glasses.} 
    \label{fig:aproach}
\end{figure}

\section{Introduction}

Artificial intelligence is generating more and more expectations. But is it really living up to those expectations? Its use is being reviewed in all areas, from natural language processing for virtual assistants, to computer vision for citizen monitoring systems or medical follow-up \cite{stone2016artificial}. Deep Neural Networks play a key role in the deployment of machine learning models in these applications. But although these algorithms achieve impressive prediction accuracies, their structure makes them very opaque. Data-driven learning processes make it difficult to control the factors and understand the information from the input data that actually drive their decisions. In this environment new efforts are being devoted to making systems more understandable and interpretable by humans \cite{mahendran2015understand,montavon2018understanding,bau2020understanding}. More concretely, new techniques are being developed to understand and visualize what machine learning models learn \cite{zeiler2014visualizing,koh2017understanding}, as well as models that generate text-based explanations of the decisions they make \cite{barredo2020xai,ortega2021xai}. 

On the other hand, thanks to adequate public outreach and debate, more and more investigations are emerging that uncover some erratic and biased behaviors of these artificial intelligence systems. These errors and biases are calling into question the safety of AI systems, both because of privacy issues \cite{2021_EncCrypto_BioSec_Fierrez} and unintended side effects \cite{serna2020discrimination}.

One way to build trust into AI systems is to relate their inner workings to human-interpretable concepts \cite{bau2020understanding}. But research is showing that not all representations in the convolutional layers of a DNN correspond to natural parts, raising the possibility of a different decomposition of the world than humans might expect, requiring further study into the exact nature of the learned representations \cite{yosinski2015visualization,geirhos2019bias}.


In this regard, bias detection is a major challenge to ensure trust in machine learning and its applications \cite{ntoutsi2020bias,terhorst2021biases}. Recent approaches for bias detection focus on the analysis of model outcomes or the visualization of learned features at the data input level \cite{zisserman2018BlindEye,zhang2018examining}. That is, they are data-bound and need inference to gain insight (see Fig. \ref{fig:aproach}). We propose a novel approach focused solely on what the Neural Networks learn (i.e., the weights of the network), freeing our method from the pitfalls of possible conflated biases in the considered datasets used for inference. The main contributions of this work can be summarized as:

\begin{itemize}
    \item We propose IFBiD, a novel bias detector trained with weights of biased and unbiased learned models. 
    \item We analyze how bias is encoded in the weights\footnote{We used the terms parameters and weights indistinctly to refer the learned filters of a Neural Network.} of deep networks through two different \textit{Case Studies} in image recognition: \textit{A}) digit classification, and \textit{B}) gender detection from face biometrics.  
    \item Our results demonstrate that bias can be detected in the learned weights of Neural Networks. This work opens a new research line to improve the transparency of these algorithms.
    \item We present two novel databases composed by 84K models trained with different types of biases. These databases are unique in the field and can be used to further research on bias analysis in machine learning.
   
\end{itemize}

\section{Related Works}

To the best of our knowledge there are no prior works that have attempted to detect the bias of a network by modeling it from learned weights. Existing literature in bias analysis focuses on the performance (outcome) \cite{bolukbasi2016word,zisserman2018BlindEye,geirhos2019bias,lieyang2019bias} and those focused on learned representations are few \cite{stock2018bias,serna2020insidebias}.

\subsection{Bias Explainability}

There is significant work on understanding neural networks learning processes, which has been useful for diagnosing CNN representations to gain a deep understanding of the biased features encoded in a CNN. In general, they map an abstract concept (e.g. a predicted class) into a domain that the human can make sense of, e.g. images or text; or they collect features of the interpretable domain that have contributed for a given example to produce a decision.

When an attribute often appears alongside other specific visual features in training images, the CNN may use these features to represent the attribute. Thus, features that appear together, but are not semantically related to the target attribute, are considered biased representations. \citet{zhang2018examining} presented a method to discover such potentially biased representations of a CNN. 

\citet{nagpal2019Face} used Class Activation Maps (CAMs \cite{zhou2016CAM}) to obtain the most discriminative regions of interest for input face images in deep face recognition models, and observed that activation maps vary significantly across races.

Also, some investigations show how psychology-inspired approaches can help elucidate bias in DNNs. Examples include \citet{ritter2017shapebias,geirhos2019bias}, who found that CNNs trained on ImageNet exhibit a strong bias towards recognizing textures rather than shapes or color. This contrasted sharply with evidence from human behavior, and revealed fundamentally different classification strategies between CNNs and humans.




\begin{figure*}[t]
    \centering
    \includegraphics[width=\textwidth]{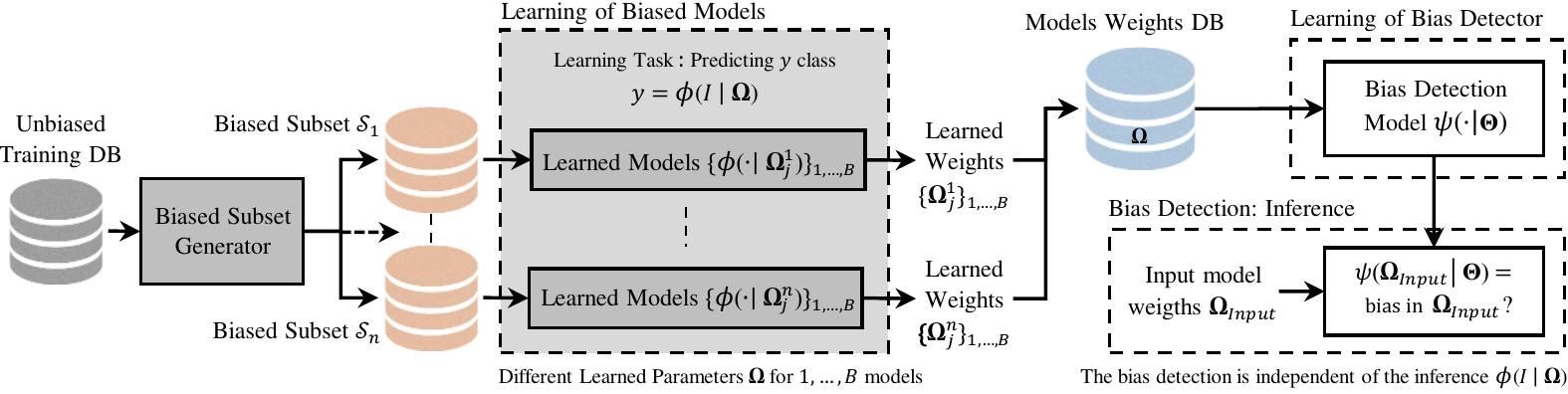}
    \caption{Learning framework of the IFBiD approach based on learned Neural Networks weights.}
    \label{fig:block_diagram}
\end{figure*}

\subsection{Bias Detection}

Research on bias analysis focuses largely on detecting causal connections between attributes in the input data and outcomes of the learned models \cite{balakrishnan2021towards}. This kind of research relies primarily on observational studies where the main conclusions are drawn from benchmarking the learned models. However, in real life applications, it is highly difficult to measure the impact of different covariates on the outcome of a learned model (i.e., it is necessary to demonstrate that correlation implies causation). \citet{balakrishnan2021towards} proposed the use of Generative Models to develop causal benchmarks applied to face analysis algorithms. These Generative Models allow manipulation of attributes in the input data, but as the authors mentioned, the synthesis methods are far from being fully controllable and there are still hidden confounders to be considered in these benchmarks. 

\citet{stock2018bias} used an adversarial example approach to model critique \cite{kim2016interpretability} by feeding the model with a carefully hand-selected subset of examples to subsequently determine whether or not it is biased.
\citet{schaaf2021towards} introduced different metrics to reliably measure several attribution maps’ techniques (Grad-CAM, Score-CAM, Integrated Gradients, and LRP-$\epsilon$) capability to detect data biases. 
\citet{gluge2020not} attempted to quantify racial bias by clustering the embeddings obtained from the model, but observed no correlation between separation in embedding space and bias.


Our work goes beyond proposals that seek to model bias through the observation of the model outcome in response to particular inputs. \citet{adebayo2018sanity} already reported the inconsistency of some widely deployed saliency methods, as they are not independent of the data. The present work follows a similar strategy to \citet{serna2020insidebias}, who uses the information learned from the model to discover bias by observing the activation of neurons to particular attributes in the inputs. The present work, however, relies solely on the information encoded in the model, without looking at particular input/outputs of the model, thus in an Inference-Free way, with the significant benefits that this represents with respect to all previous works.

The hypothesis behind our proposed Inference-Free Bias Detection (IFBiD) is that bias is encoded in the parameters of a learned model and it can be detected. IFBiD is an interesting and noteworthy effort to contribute to tackling the bias problem. The inference of IFBiD is performed directly over the weights of a learned model, and therefore does not require a causal benchmark based on input/output analysis.

\section{Problem Statement and Proposed Approach}

In this work we adopt the formulation proposed by \citet{kleinberg2019discrimination} and \citet{serna2020discrimination} for the discrimination of groups of people, but applicable to any type of bias. The formalization follows:

\textbf{Definition 1} (Data). $\mathcal{D}$ is a dataset (collection of multiple samples from different classes) used for training and/or evaluating a model $\mathcal{M}$. Samples in $\mathcal{D}$ can be classified according to some criterion $\textit{d}$. The set $\mathcal{D}_d^c \subset \mathcal{D}$ represents all the samples corresponding to class $c$ of criterion $d$. 

\textbf{Definition 2} (Learned Model). The learned model $\mathcal{M}$ is trained according to input data $\mathcal{I} \subset \mathcal{D}$, a Target function $T$ (e.g., digit classification or gender detection), and a learning strategy that maximizes a goodness criterion $G$ on that task (e.g., typically a performance function) based on the output $O$ of the model and the Target function $T$ for the input data $\mathcal{I}$. 

\textbf{Definition 3} (Biased Model). A learned model $\mathcal{M}$ is biased with respect to a specific class $c$ of criterion $d$ if the goodness $G$ on task $T$ when considering the full set of data $\mathcal{D}$ is significantly different to the goodness $G(\mathcal{D}_d^c)$ on the subset of data corresponding to class $c$ of the criterion $d$.

Typically, as in our case, the model $\mathcal{M}$ is a Neural Network $\phi(\cdot)$, parameterized by $\mathbf{\Omega}$, and the goodness-of-fit criterion consists in minimizing an objective function (e.g. the cross-entropy loss function).


The training process of Neural Networks is usually not deterministic and the resulting parameters $\mathbf{\Omega}$ depend on several elements: training data, learning architecture (e.g., number of layers, number of neurons per layer, etc.), training hyper-parameters (e.g., loss function, number of epochs, batch size, learning rate, etc.), initialization parameters, and optimization algorithm.

The existing literature on bias analysis is mainly focused on the inputs $\mathcal{I}$ \cite{tommasi2017bias,zhang2018examining,wang2020bias} and the outputs $O$ to given inputs \cite{buolamwini2018GenderShades,zisserman2018BlindEye,serna2020discrimination}. We propose a novel approach to detect bias in the learned parameters $\mathbf{\Omega}$, regardless of the particular input $\mathcal{I}$ or the output $O=\phi(\mathcal{I}|\mathbf{\Omega})$ (see Fig. \ref{fig:block_diagram}).

\begin{figure*}[t]
    \centering
    \includegraphics[width=\textwidth]{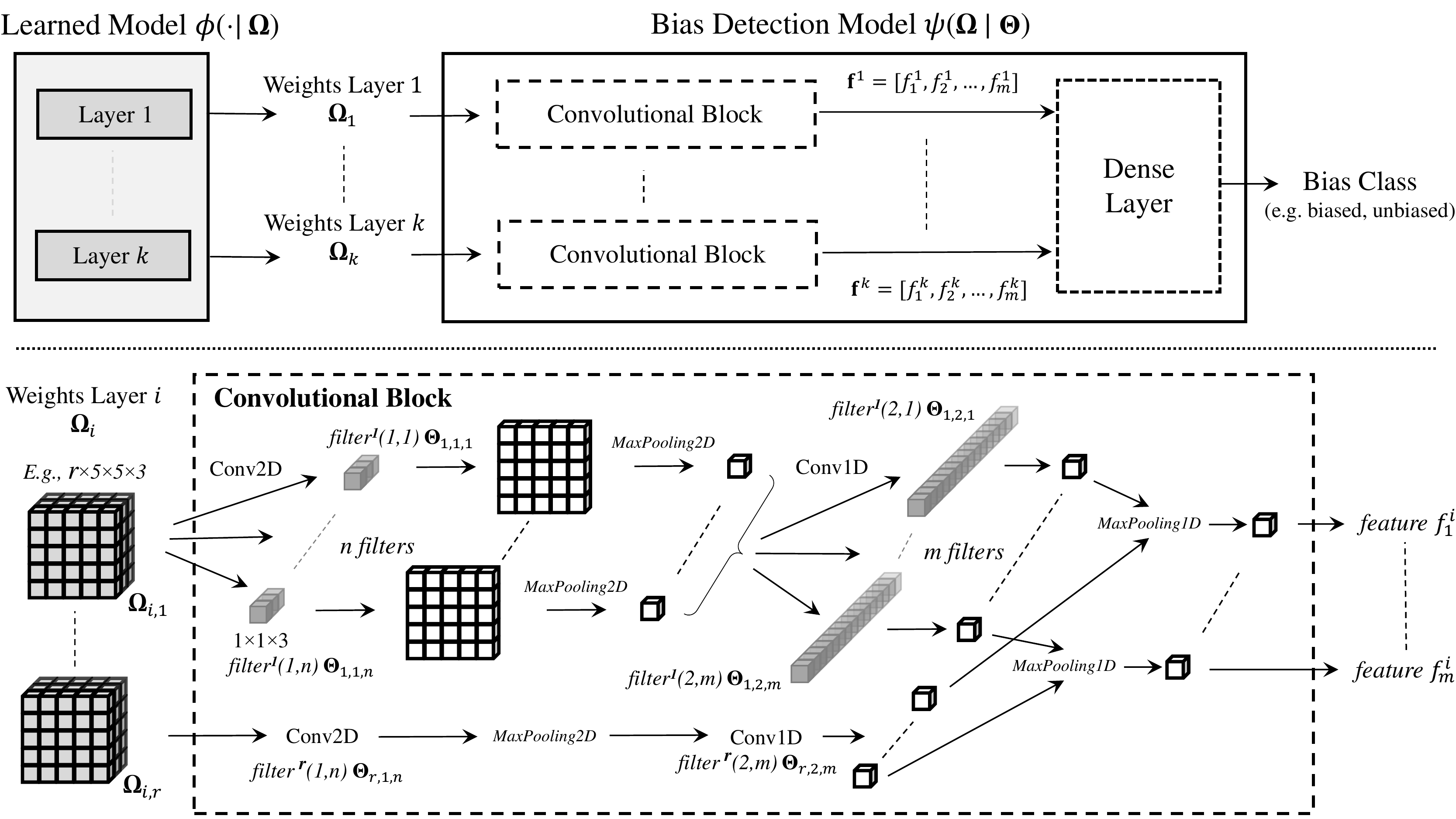}
    \caption{General architecture of a bias detector with the $1\times1$-conv module variant. The architecture depends on the number of layers $k$ of the model $\mathbf{\Omega}$ to be audited. The depth of the module filters depends on the depth of the input weights. Module variant $1\times1$-conv consists of the subsequent layers: $1\times1$ convolution followed by $d\times d$ MaxPooling, then again a one-dimensional convolution with kernel size of $1$ followed by a MaxPooling with pool size equal to the number of input filters. Do not confuse the suffixes in this figure ($k$,$i$ indicates layer) with those in Fig. \ref{fig:block_diagram} ($j$ indicates model number). }
    \label{fig:architecture}
\end{figure*}

\subsection{IFBiD: Inference-Free Bias Detection Learning}

The aim of the bias detection model is to find patterns in $\mathbf{\Omega}$ associated with biased outcomes. We designed the bias detector as a Neural Network $\psi(\cdot)$ represented by its parameters $\mathbf{\Theta}$.

In our approach (detailed in next sections), we train the bias detector using a dataset of biased and unbiased models. The models $\phi(\cdot |\mathbf{\Omega})$ for task $T$, are biased by training them with biased subsets of the database $\mathcal{S}_1, ..., \mathcal{S}_n \subset \mathcal{D}$. To build a training set for the detector $\psi(\cdot |\mathbf{\Theta})$, we train a number of models $\phi(\cdot |\mathbf{\Omega})$ with each subset $S_i$, forming $\{\mathbf{\Omega}^i_j\}^{i=1,...,n}_{j=1,...,B}$ (see Fig. \ref{fig:block_diagram}), where $n$ is the number of biased subsets and $B$ is the number of models trained with each biased subset. $\mathbf{\Omega}^{i}_{j}$ denotes the $j$th instance of a learned model trained with biased subset $i$. 

Because of the non-deterministic nature of the training process of the network $\phi(\cdot |\mathbf{\Omega})$, the same training subset $i$ is likely to give rise to different $\mathbf{\Omega}$ (i.e., $\mathbf{\Omega}^{i}_{j} \neq \mathbf{\Omega}^{i}_{k}$). The reason for this is that since the solution space is very large, the solution (which is iteratively approximated) typically arrives at a local minimum that depends on the initialization, the particular training configuration, and the order of the data \cite{lecun2015deep}. In CNNs, this translates into the fact that filters tend to differ between networks. The bias detector, therefore, has to be able to detect similar filters in different positions and configurations. The problem is analogous to detecting patterns in images, where one can be in different parts of an image. 

It is important to underline that the approach requires the target DNNs (i.e., $\Omega_{Input}$ in Fig. 2) to have exactly the same architecture as the DNNs used to train the detector. This does not detract from the fact that the task is still challenging, since, as we have explained before, the filters learned by a convolutional network never appear in the same place and are never identical due to the randomness of data presentation and weight initialization (which is, incidentally, the reason why we have also used convolutions in the detector architecture).

\subsection{The Detector}

We evaluated many different learning architectures for IFBiD. The design of the possible architectures has not only taken into account the number and types of layers, but has also depended on the selection of the parameters $\mathbf{\Omega}$ of the model $\phi(\cdot |\mathbf{\Omega})$ used as input for the detector $\psi(\cdot |\mathbf{\Theta})$.

The detector architecture consists of a module to process the weights/filters of each layer, and then a dense layer that concatenates all the outputs of each module. The bias detector architecture consists of multiple modules to process the weights/filters of each layer (thus one module for each layer), and then a fully connected layer that concatenates all the outputs of each module. Fig. \ref{fig:architecture} shows the general architecture designed for a specific module variant (see below). The components of a module are the same for all layers (convolution, maxpooling, etc.), as well as their order; the only thing that changes are their parameters, which depend on the size of the input weights.

We have developed different approaches, in which the general architecture remains stable, and what changes are the modules. The module variants we analyzed were the following (where $d\times d$ is the dimension of the input filter weights, $c$ is the number of input channels, and $r$ is the number of input filters):

\begin{itemize}
    \item \textbf{MLP}: Flatten $\rightarrow$ Dense($r$)
    \item \textbf{\boldmath$1\times1$ +conv}: Conv2D ($1\times1$) $\rightarrow$ MaxPooling2D ($d\times d$) $\rightarrow$ Conv1D (1) $\rightarrow$ MaxPooling1D ($r$) $\rightarrow$ Flatten. 
    \item \textbf{\boldmath$1\times1$ +max}: Conv2D ($1\times1$) $\rightarrow$ MaxPooling3D ($d\times d\times k$) $\rightarrow$ Flatten.
    \item \textbf{\boldmath$1\times1\times1$ +max}: Conv3D ($1\times1\times1$) $\rightarrow$ MaxPooling3D ($d\times d \times c$) $\rightarrow$ Flatten.
\end{itemize}

Convolutions are followed by a relu activation function, and there is always $0.1$ dropout afterwards (we have seen that it works best among the values: 0.0, 0.1, 0.2 and 0.3).

\section{Experiments}

\subsection{Datasets of Biased Models}

We have created two databases for experimenting in automatic bias detection: DigitWdb and GenderWdb. The databases contain the weights $\mathbf{\Omega}$ of the models $\phi(\cdot, |\mathbf{\Omega})$ used in our experiments for the tasks of digit and gender classification. The databases include $84$K models trained with different types of biases (each model has an associated label identifying the bias). These databases are publicly available for further research.\footnote{\url{https://github.com/BiDAlab/IFBiD/}}

\subsubsection{Case Study A: Digit Classifier (DigitWdb).}
\label{subsubsec:DigitDB}

We have put together a database that contains the weights $\mathbf{\Omega}$ of $48$K  digit classification networks $\phi(\cdot, |\mathbf{\Omega})$. For this we have used the colored MNIST database \cite{kim2019learning}, which consists of seven replicas of the MNIST database, each with a different level of color bias (of which we have only used four).

To synthesize the color bias, ten different colors were selected and assigned to each digit category as its mean color. Then, for each training image, a color was randomly sampled from the normal distribution of the corresponding mean color, and the digit was colorized. The level of bias of each replica depends on the value of the variance used in the normal distribution: the lower the more bias.

The architecture is the same for all models: a CNN $\phi(\cdot |\mathbf{\Omega})$ with three convolutional layers with relu activation, each followed by a maxpool, and two fully connected layers at the end (with $128$ and $10$ neurons, a relu and a softmax activation function respectively), with a dropout layer of 0.3 between the two. Each of the trained models results in a total of $50$K parameters.

All model parameters $\mathbf{\Omega}$ have been initialized randomly with Glorot uniform \cite{glorot2010init} to avoid possible commonalities. A diagram showing the general construction of a weight database is shown in Fig. \ref{fig:block_diagram}. The composition is as follows:

\begin{itemize}
    \item Train: $40$K models classified by bias level into four groups, with $10$K models per level ($B=10$K). The models were trained using the first $30$K training digits from Colored MNIST. The models have been categorized into four groups depending on the replica subset with which they have been trained ($n=4$). The level of bias of the replica subset is what determines the level of bias of the model.  Groups are: very high bias (color jitter variance of $0.02$), high bias (color jitter variance of $0.03$), low bias (color jitter variance of $0.04$), and very low bias (color jitter variance of $0.05$).
    
    \item Test: $8$K models classified by bias level into four groups ($2$K models for each level). The models were trained using the last $30$K training digits from Colored MNIST and categorized in the same way as the training ones (i.e., from very high bias to very low bias). 
       
\end{itemize}

Each Colored MNIST biased subset has $60$K training digits, so the $30$K for train and $30$K for test are independent. This means that the DigitWdb models assigned to test have learned with different data than the DigitWdb models assigned to train.

\textbf{Properties}. All models have the same architecture and similar class performance. 
In this case study, the bias is determined by the color jitter variance of the digit images.

\begin{table}
\normalsize
  \begin{center}
    \caption{Average digit classification accuracy (in \%) in DigitWdb models according to their level of bias. Classification accuracy has been assessed with the Colored MNIST test set, a set of randomly colored numbers, and therefore not biased.}\smallskip
    \label{table:digits_acc}
    \resizebox{1\columnwidth}{!}{ 
    \begin{tabular}{ccccccccccc}
      \toprule
      \multirow{2}{*}{\textbf{\shortstack{Model\\Bias}}} & \multicolumn{10}{c}{\textbf{Digit Classification Accuracy}}
      \tabularnewline
      \cmidrule(lr){2-11}
      & \textbf{0}  & \textbf{1} & \textbf{2} & \textbf{3} & \textbf{4} & \textbf{5} & \textbf{6} & \textbf{7} & \textbf{8} & \textbf{9}
      \tabularnewline
      \cmidrule(r){1-1}\cmidrule(lr){2-2}\cmidrule(lr){3-3}\cmidrule(lr){4-4}\cmidrule(lr){5-5}   \cmidrule(r){6-6}\cmidrule(lr){7-7}\cmidrule(lr){8-8}\cmidrule(lr){9-9}\cmidrule(lr){10-10}\cmidrule(lr){11-11}
        Very Low	&	88 & 94 & 77 & 82 & 90 & 89 & 75 & 84 & 81 & 82
        \tabularnewline     									
        Low	&	79  &  85  &  67  &  69  &  81  &  83  &  65  &  76  &  72  &  69
        \tabularnewline     									
        High	&	66  &  76  &  54  &  56  &  72  &  69  &  49  &  64  &  58  &  46 
        \tabularnewline     									
        Very High	&	49 &  51 &  42 &  38 &  59 &  40 &  40 &  51 &  43 &  32  
        \tabularnewline
      \bottomrule
    \end{tabular}
    }
  \end{center}
\end{table}

Table \ref{table:digits_acc} shows the average digit classification accuracy of all models trained with the four subsets of different bias levels, from very low bias (subset $\mathcal{S}_1$ with variance of $0.05$) to very high bias (subset $\mathcal{S}_4$ with variance of $0.02$). The table shows performance on the Colored MNIST's test set, that is, a set of randomly colored numbers, and therefore not biased.

Although not shown, all models exceeded 99\% accuracy during training (i.e. in their respective training subsets $\mathcal{S}_i$). This means that our models learn as far as the training set allows. But then, in the (unbiased) test set (i.e., with all digits colored randomly) the number of correct classifications drops considerably. The reason is that the color (present in the training set in a biased way) has been learned as a differentiating element when classifying digits. Thus, the network was not only learning to associate a number to a shape, but also to a color. This is why, subsequently, when finding a digit with a random color, it has much more difficulty in classifying it correctly.

Also, Table \ref{table:digits_acc} shows a clear difference in the performance of the models as a function of the level of bias of the dataset with which it has been trained. This difference is the basis for the experiments carried out in this work.


\subsubsection{Case Study B: Gender Classifier (GenderWdb).} \label{sec:GenderDB}

\begin{table}
\normalsize
  \begin{center}
    \caption{Average gender classification accuracy (\%) in gender classification of all GenderWdb models, according to their class bias. Models are trained with DiveFace dataset.}\smallskip
    \label{table:gender_acc}
    \begin{tabular}{lccc}
      \toprule
      \multirow{2}{*}{\textbf{Model Bias}} & \multicolumn{3}{c}{\textbf{Gender Classification Accuracy}}
      \tabularnewline
      \cmidrule(lr){2-4}
      & \textbf{Asian} & \textbf{Black} & \textbf{Caucasian}
      \tabularnewline
      \cmidrule(r){1-1}\cmidrule(lr){2-2}\cmidrule(lr){3-3}\cmidrule(lr){4-4}
		Asian	&	89.5\%	&	81.5\% & 82.9\%
		\tabularnewline     
		Black	&	82.0\%	& 89.4\% & 83.0\%
		\tabularnewline     
		Caucasian	&	80.0\%	& 83.2\% & 89.2\%
		\tabularnewline     
	\bottomrule
    \end{tabular}
  \end{center}
\end{table}

We have gathered together a database that contains the weights $\mathbf{\Omega}$ of $36$K gender classification networks $\phi(\cdot |\mathbf{\Omega})$. We are aware there are more gender categories other than male and female. Since establishing ground-truth genetic sex is not possible, we use gender as a proxy for sex. We use it as a simplified application of a real-life based problem.

To train the gender classification models that constitute this database, we used DiveFace \cite{SensitiveNets2021}. DiveFace is a face dataset contaning $24$K identities and three images per identity. Identities are evenly distributed according to gender: male and female, and three categories related to ethnic physical characteristics: Asia, African/Indian, and Caucasian.

As in the DigitWdb database, we have separated the data for training into two independent sets of the same size. (This serves to ensure the future independence of the bias detector training and testing.) 

We have trained $36$K gender classification models $\phi(\cdot |\mathbf{\Omega})$, divided into $30$K for training and $6$K for testing. The architecture is the same for all models: a CNN with six convolutional layers with relu activation, each followed by a maxpool, and two fully connected layers at the end (with $128$ and two neurons, a relu and a softmax activation function respectively). The result is a model with a total of $100$K parameters.

All model parameters $\mathbf{\Omega}$ have been initialized randomly with Glorot uniform \cite{glorot2010init} to avoid possible commonalities. The composition is as follows:

\begin{itemize}
    \item Train: $30$K models belonging to three classes of bias ($n=3$), depending on the subset $\mathcal{S}_i$ with which the model has been trained, with $10K$ models per class ($B=10$K). The models were trained using the first $12$K faces of each ethnic group of DiveFace: $\mathcal{S}_1$ is asian biased, $\mathcal{S}_2$ is black biased, and  $\mathcal{S}_3$ is caucasian biased.
    
    \item Test: $6$K models with the same three types of bias as train. The models were trained using the last $12$K faces of each ethnic group of DiveFace.
      
\end{itemize}

\textbf{Properties}. All models have the same architecture and similar class performance. Table \ref{table:gender_acc} shows the average accuracy in gender classification of all models, separated by bias. Bias has been introduced through the subset $\mathcal{S}_i$ with which the model has been trained. In this case study the bias is determined by the ethnicity of the face images. What becomes clear from looking at the table is the strong bias in the performance of the models in each of the groups. Note that ethnicity attributes include the color of the skin, but also more complex anthropomorphic face features.

\subsection{Results}

\textbf{Bias in digit classification models (Case Study A)}. First of all we have attempted a binary classification problem: detect strong bias against minimal or no bias. 
\begin{figure}[t]
    \centering
    \includegraphics[width=\columnwidth]{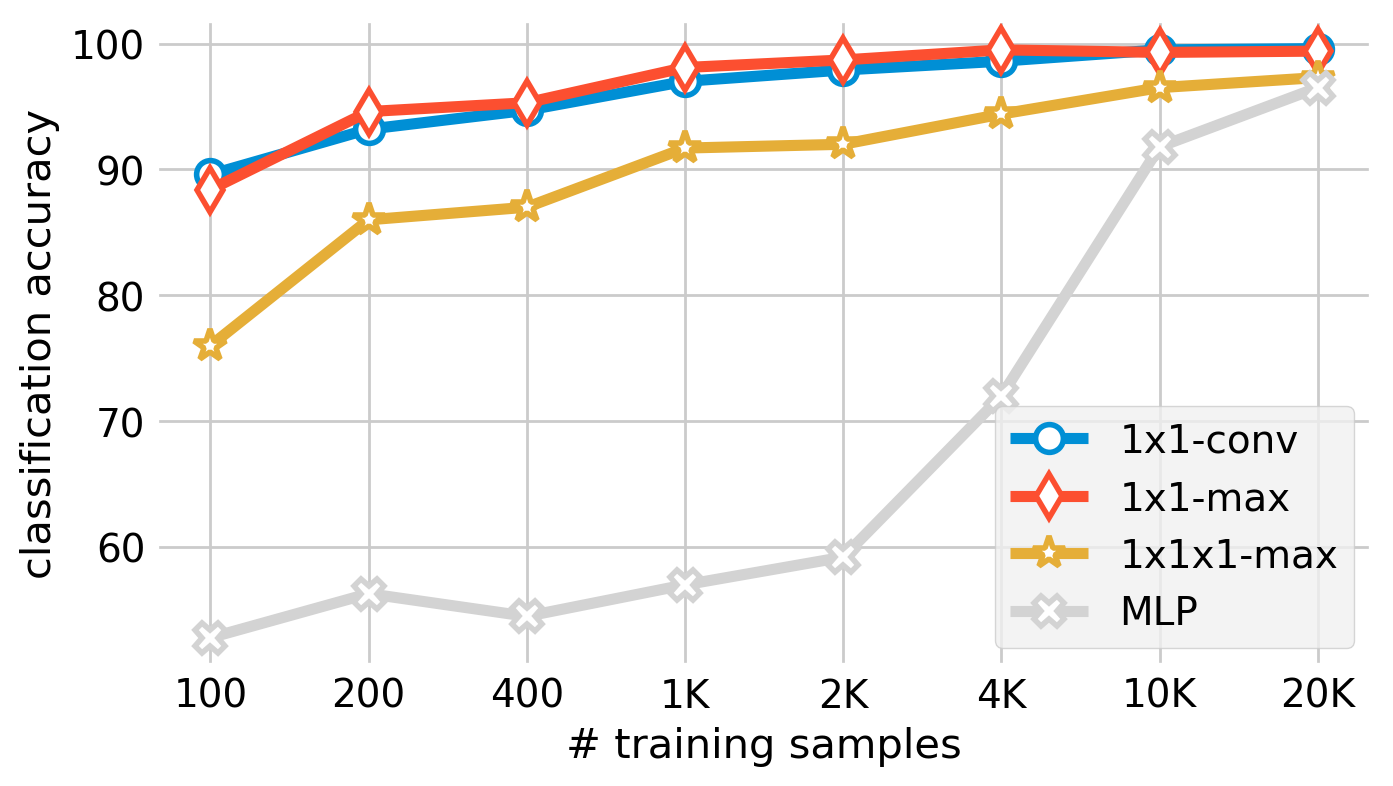}
    \caption{Bias detection accuracy in DigitWdb for the different architectures given the number of training samples (x axis).}
    \label{fig:digit_binary}
\end{figure}

In this first case we have used models with very high bias and very low bias. Fig. \ref{fig:digit_binary} shows the accuracy of bias detection in digit classification models $\phi(\cdot |\mathbf{\Omega})$ for the different architectures given the number of samples the detector $\psi(\cdot |\mathbf{\Theta})$ was trained with. It can be seen that the convolutional architectures show a saturation of classification performance and that it does not take many samples to get great performance. In fact, with the best architectures, $100$ training samples are sufficient to achieve a performance of around $90\%$. These initial results suggest that bias is encoded in the weights $\mathbf{\Omega}$ of the learned models $\phi(\cdot)$ and it can be detected.

A second experiment has been trying to detect the level of bias of a model $\phi(\cdot)$, or in other words, to classify the models according to their level of bias. This is a more complex problem and has required us to test more architectures for the detector $\psi(\cdot)$.

\begin{figure}[t]
    \centering
    \includegraphics[width=\columnwidth]{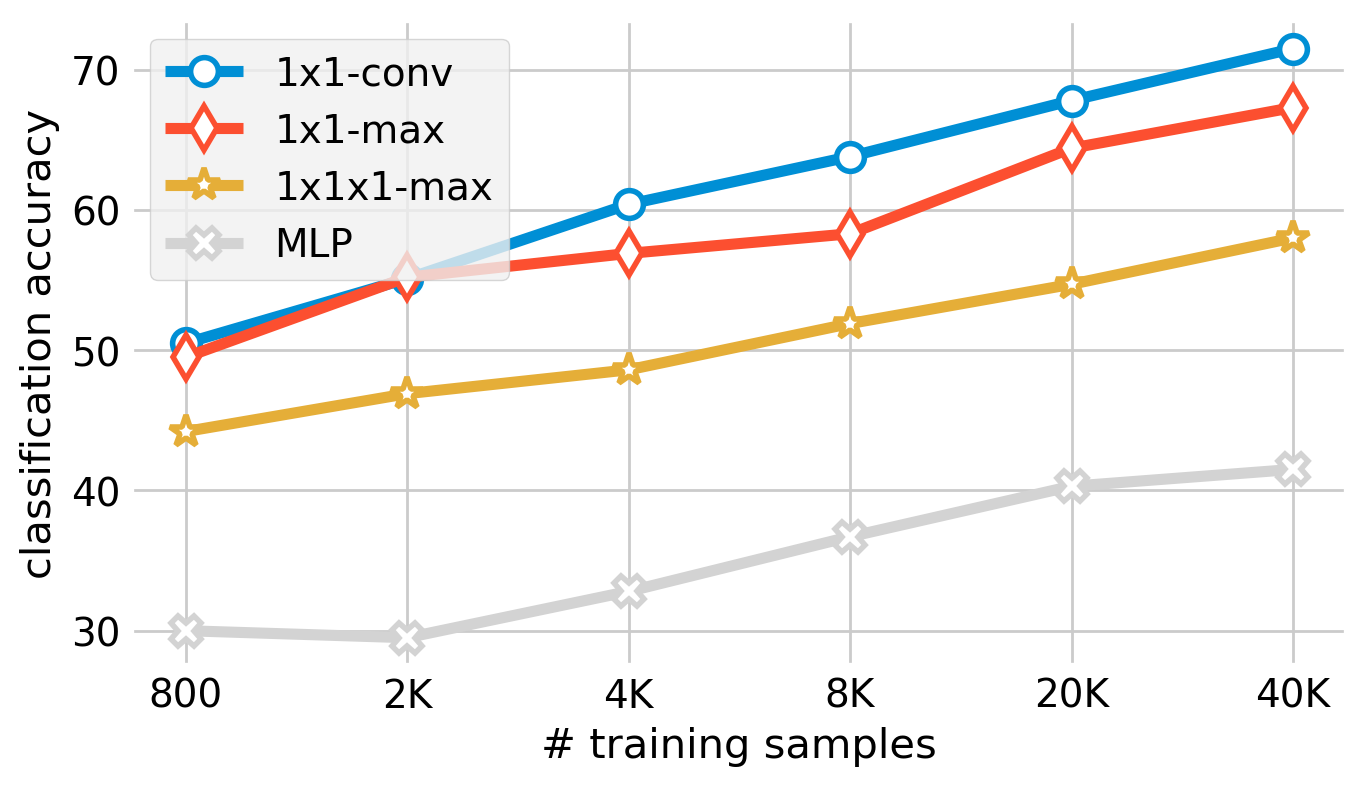}
    \caption{Classification accuracy of the bias level in DigitWdb for the different architectures given the number of training samples (x axis).}
    \label{fig:digit_four}
\end{figure}

Fig. \ref{fig:digit_four} shows the classification accuracy of the $4$ bias levels (cf. initial subsection within Experiments describing the Datasets for Case Study A) for the digit classification models $\phi(\cdot |\mathbf{\Omega})$ and the different architectures given the number of samples with which the detector $\psi(\cdot |\mathbf{\Theta})$ was trained. We see that distinguishing the level of bias in digit classification models is more complicated than simply stating bias-no bias, and that in this case the maximum success rate we achieve in the classification is $70\%$ (note that random chance is $25\%$ for this task). Another important thing to note is the tendency (of good architectures) to keep improving as the training set is increased, they do not seem to be reaching their performance limit. 

\textbf{Bias in gender classification models (Case Study B)}. After seeing positive results, we made the leap to a more complex problem (i.e., more covariates): detecting ethnic bias in models trained for gender recognition.

\begin{figure}[t]
    \centering
    \includegraphics[width=\columnwidth]{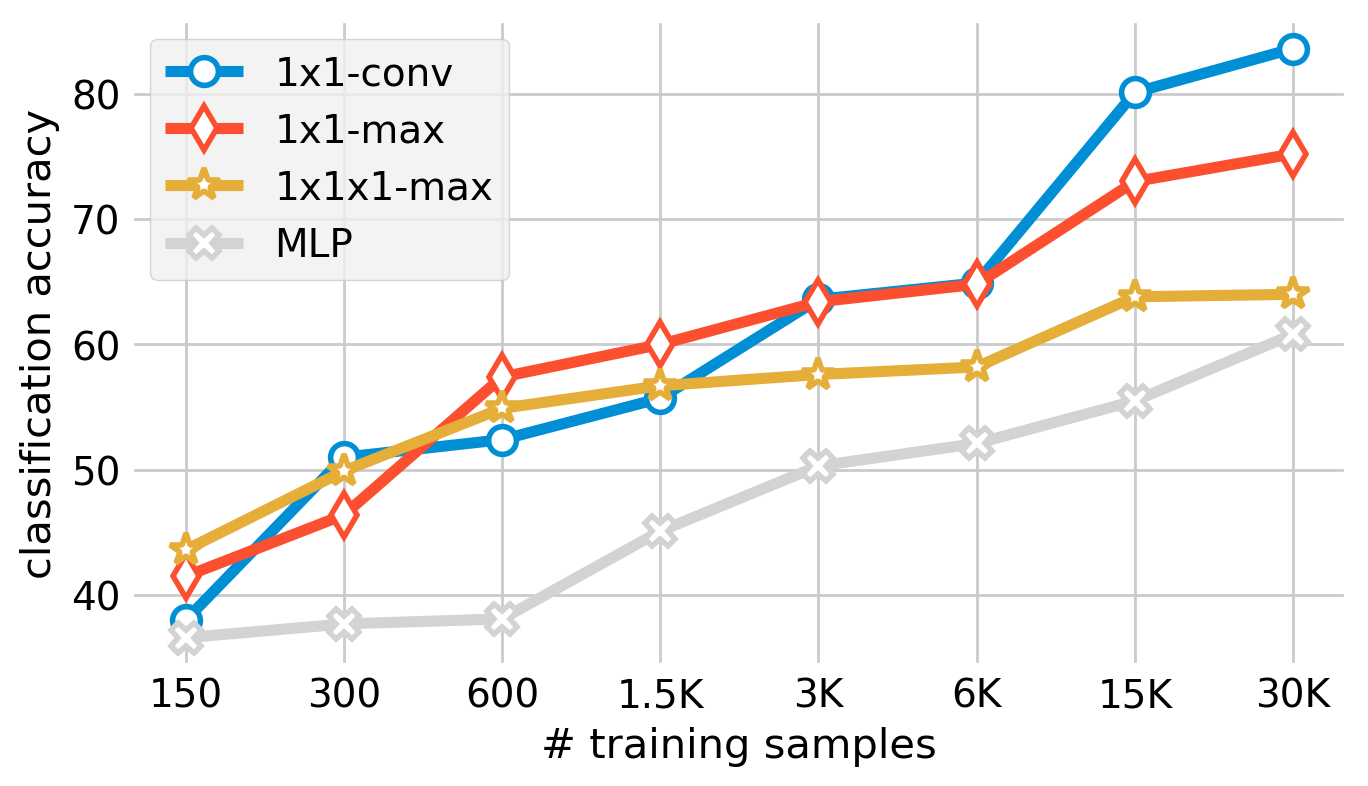}
    \caption{Bias classification accuracy in GenderWdb, for the different architectures given the number of training samples (x axis). Bias is classified into three different categories according to a ethno-demographic criterion, namely Asian, Black, and Caucasian.}
    \label{fig:gender_detect}
\end{figure}

Fig. \ref{fig:gender_detect} shows the bias classification accuracy of the biased gender recognition models $\phi(\cdot)$ for the different architectures and the number of samples used for training.

The curves show that after a certain number of training samples, the accuracy is no longer increasing in the model with more parameters (that containing Conv3D: $1 \times 1 \times 1$-max). However, it can be seen that when trained with little data, it performs similarly to the rest. The hypothesis that best seems to explain this behavior is that, since there are so many parameters $\mathbf{\Theta}$, the solution space is so large that the choice of a better architectural configuration occurs automatically, leaving unnecessary parameters unchanged, as if they were not present \cite{schmidt1992weights}. With little training data the model $\psi(\cdot | \mathbf{\Theta})$ adjusts very quickly to those data (losses are practically nil) and in just a couple of epochs it no longer needs to adjust those weights $\mathbf{\Theta}$. On the other hand, when the number of training samples increases, it needs to modify more parameters in order to correlate the training data well, thus losing the generalization capability equivalent to architectures with fewer parameters.

The gender classifier $\phi(\cdot)$ has more layers than the digit classifier $\phi(\cdot)$, twice as many. So the bias detection network $\psi(\cdot)$ for these models has more parameters, and thus the performance is different. The best performance is obtained with the same architecture that also obtains the best performance in the digit models, the architecture with two convolutions: $1\times 1$-conv; reaching $83\%$ detection accuracy. The improvement in the MLP, that has the most parameters, seems to be growing steadily. The rest of the architectures seems that from 15k training samples onwards is when doubling the samples does not increase its performance so much. But it would be necessary to keep doubling the number to check if the trend holds.


We have dealt with many more architectures that are not worth describing here: using two dense layers at the end, adding a dense layer after each convolution, replacing convolutions with dense layers, playing with dropout, etc.; all resulting in worse performance than the learning architectures reported here.



\subsection{SOTA Comparison}

Table \ref{table:sota} shows the comparison with a recent state-of-the-art bias detection method \cite{serna2020insidebias}, which consists of measuring the activation of the last layer upon image input. We also use an SVM with radial basis function (RBF) kernel as a baseline, trained in the same way as our detector. The table shows the percentage of biased models detected by InsideBias, the RBF SVM, and by our method. The 6K GenderWdb test models were used, being 2K of each type of bias. 

In order to apply InsideBias, we used 60 images from DiveFace, with 20 of each ethnicity and the same number of men and women. Separately, as input to the SVM, we used the parameters of the models put together as a vector of length 97K. 

Our method shows considerable superiority: it has a good hit performance on models with all biases, whereas the other methods only detect well a single type of bias in the models.

\begin{table}[t]
\normalsize
  \begin{center}
    \caption{Percentage of biased models detected according to their type of bias.}\smallskip
    \label{table:sota}
    \resizebox{1\columnwidth}{!}{
    \begin{tabular}{lccc} 
      \toprule
      \multirow{2}{*}{\textbf{Method}} & \multicolumn{3}{c}{\textbf{Bias Detection Accuracy}}
      \tabularnewline
      \cmidrule(lr){2-4}
      & \textbf{Asian} & \textbf{Black} & \textbf{Caucasian}
      \tabularnewline
      \cmidrule(r){1-1}\cmidrule(lr){2-2}\cmidrule(lr){3-3}\cmidrule(lr){4-4}
      RBF SVM & 71\% &	30\% & 26\%\\ 
      InsideBias \cite{serna2020insidebias} & 23\% &	86\% & 3\%\\ 
      \textbf{IFBiD (ours)} & 95\% &	79\% & 79\%\\
      \bottomrule 
    \end{tabular}
    }
  \end{center}
\end{table}

\section{Conclusion}

We presented a novel approach called IFBiD (Inference-Free Bias Detection) to analyze biases in neural networks: by auditing the models through their weights. Our experiments demonstrate the existence of identifiable patterns associated with bias in the weights of a trained Neural Network \cite{terhorst2021biases}. We conducted experiments in two computer vision use cases: digit and face gender classification \cite{serna2020insidebias}. This involved generating two databases with thousands of biased models each. The first, DigitWdb, with models trained on the Colored MNIST database \cite{kim2019learning}; and the second, GenderWdb, with models trained on a face database, DiveFace \cite{SensitiveNets2021}.

We used each database to train bias detectors following the proposed IFBiD principles. We have evaluated a number of architectures and have found that in both cases it is possible to achieve a good performance in bias detection. In the digit classification models we were able to detect whether they presented strong or low bias with more than $99\%$ accuracy, and we were also able to classify between four levels of bias with more than $70\%$ accuracy. For the face models, we achieved $83\%$ accuracy in distinguishing between models biased towards Asian, Black, or Caucasian ethnicity. In both cases the experiments are open-ended in the absence of increasing both databases. This has been evident in the plots with the experiments carried out for different sizes of the training set.

We evaluated our approach by varying the nature of the data (i.e., digits and face images), type of architecture (i.e., number of layers, units), and optimization strategy (e.g., loss function). For future work, the generalization capabilities of our proposed bias detection approach should be studied in more depth. The training process of the biased models used for training IFBiD can be affected by hidden confounders that need to be considered.

\section{Acknowledgments}

This work has been supported by projects: TRESPASS-ETN (MSCA-ITN-2019-860813), PRIMA (MSCA-ITN-2019-860315), BIBECA (RTI2018-101248-B-I00 MINECO/FEDER), and BBforTAI (PID2021-127641OB-I00 MICINN/FEDER). I. Serna is supported by a FPI fellowship from UAM.

\bibliography{References}

\end{document}